\documentclass[conference]{IEEEtran}
\IEEEoverridecommandlockouts
\usepackage{cite}
\usepackage{amsmath,amssymb,amsfonts}
\usepackage{graphicx}
\usepackage{textcomp}
\usepackage{xcolor}

\usepackage[colorlinks,urlcolor=blue,linkcolor=blue,citecolor=blue]{hyperref}

\usepackage{color,array}

\usepackage{float}
\usepackage{subcaption}

\usepackage{algorithm}
\usepackage{algpseudocode}

\DeclareMathAlphabet\mathbfcal{OMS}{cmsy}{b}{n}
\usepackage{bm}

\usepackage{stfloats}

\def\BibTeX{{\rm B\kern-.05em{\sc i\kern-.025em b}\kern-.08em
    T\kern-.1667em\lower.7ex\hbox{E}\kern-.125emX}}
\begin{document}

\title{Adversarial Driving: Attacking End-to-End Autonomous Driving\\
\thanks{The project is supported by Offshore Robotics for Certification of Assets (ORCA) Partnership Resource Fund (PRF) on Towards the Accountable and Explainable Learning-enabled Autonomous Robotic Systems (AELARS) [EP/R026173/1].
.}
}

\makeatletter
\newcommand{\linebreakand}{%
  \end{@IEEEauthorhalign}
  \hfill\mbox{}\par
  \mbox{}\hfill\begin{@IEEEauthorhalign}
}
\makeatother

\author{
\IEEEauthorblockN{1\textsuperscript{st} Han Wu}
\IEEEauthorblockA{\textit{Computer Science} \\
\textit{The University of Exeter}\\
Exeter, the United Kingdom \\
hw630@exeter.ac.uk}
\and
\IEEEauthorblockN{2\textsuperscript{nd} Syed Yunas}
\IEEEauthorblockA{\textit{Computer Science} \\
\textit{The University of the West of England}\\
Bristol, the United Kingdom \\
syed.yunas@uwe.ac.uk}
\and
\IEEEauthorblockN{3\textsuperscript{rd} Sareh Rowlands}
\IEEEauthorblockA{\textit{Computer Science} \\
\textit{The University of Exeter}\\
Exeter, the United Kingdom \\
s.rowlands@exeter.ac.uk} \\
\linebreakand
\IEEEauthorblockN{4\textsuperscript{th} Wenjie Ruan}
\IEEEauthorblockA{\textit{Computer Science} \\
\textit{The University of Exeter}\\
Exeter, the United Kingdom \\
w.ruan@exeter.ac.uk}
\and
\IEEEauthorblockN{5\textsuperscript{th} Johan Wahlstr\"om$^{*}$}
\IEEEauthorblockA{\textit{Computer Science} \\
\textit{The University of Exeter}\\
Exeter, the United Kingdom \\
j.wahlstrom@exeter.ac.uk}
}

\maketitle

\begin{abstract}
As research in deep neural networks advances, deep convolutional networks become promising for autonomous driving tasks. In particular, there is an emerging trend of employing end-to-end neural network models for autonomous driving. However, previous research has shown that deep neural network classifiers are vulnerable to adversarial attacks. While for regression tasks, the effect of adversarial attacks is not as well understood. In this research, we devise two white-box targeted attacks against end-to-end autonomous driving models. Our attacks manipulate the behavior of the autonomous driving system by perturbing the input image. In an average of 800 attacks with the same attack strength (epsilon=1), the image-specific and image-agnostic attack deviates the steering angle from the original output by 0.478 and 0.111, respectively, which is much stronger than random noises that only perturbs the steering angle by 0.002 (The steering angle ranges from [-1, 1]). Both attacks can be initiated in real-time on CPUs without employing GPUs. Demo video: \href{https://youtu.be/I0i8uN2oOP0}{https://youtu.be/I0i8uN2oOP0}.
\end{abstract}

% The driving system uses a regression model that takes an image as input and outputs the steering angle. 


\section{Introduction}

Autonomous driving is one of the most challenging tasks in safety-critical robotic applications. Most real-world autonomous vehicles employ modular systems that divide the driving task into smaller subtasks. In addition to a perception module that relies on deep learning models to locate and classify objects in the environment, modular systems also include localization, prediction, planning, and control modules. However, researchers are also exploring the potential of end-to-end driving systems. An end-to-end driving system is a monolithic module that directly maps the input to the output, often using deep neural networks. For example, the NVIDIA end-to-end driving model \cite{bojarski2016end} maps raw pixels from the front-facing camera to steering commands. The development of end-to-end driving systems has been facilitated by recent advances in high-performance GPUs and the development of photo-realistic driving simulators, such as the Carla Simulator \cite{Dosovitskiy17} and the Microsoft Airsim Simulator \cite{airsim2017fsr}.  

\begin{figure}[tbp]
    \centering
    \includegraphics[width=0.48 \textwidth]{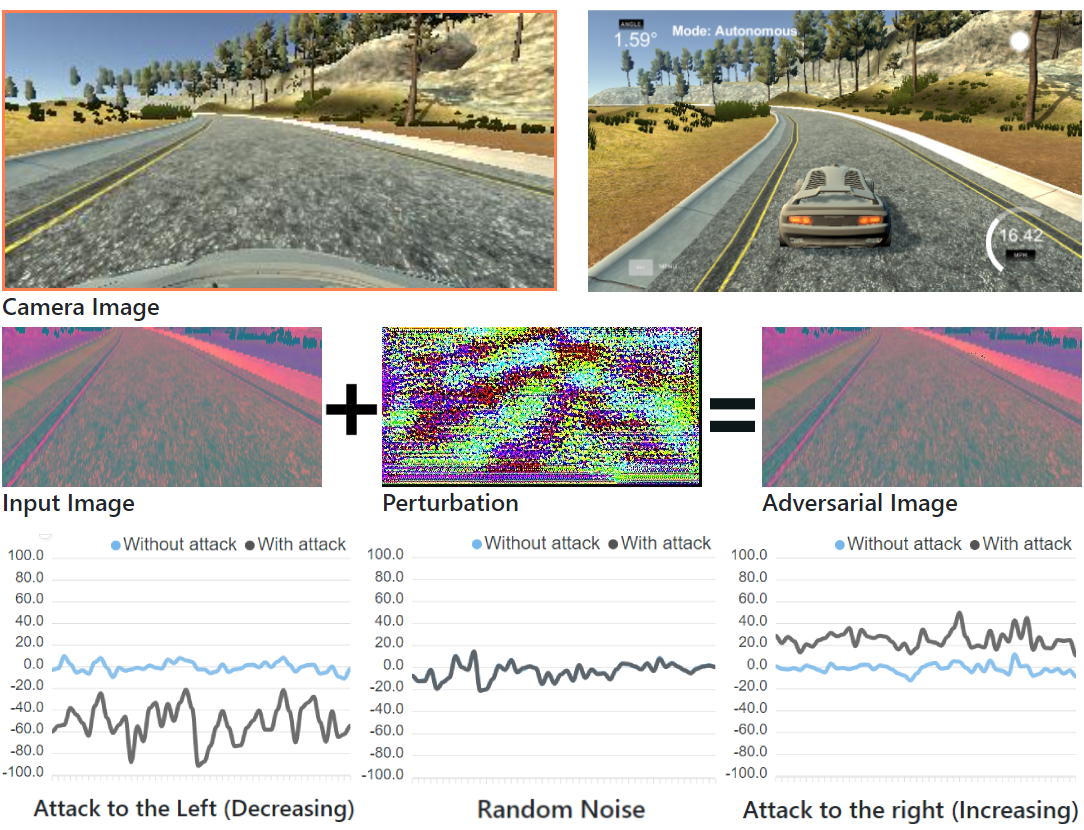}
    \caption{Adversarial Driving: The behavior of and end-to-end autonomous driving model can be manipulated by adding unperceivable perturbations to the input image.}
    \label{fig:adv_drv}
\end{figure} 

As demonstrated in multiple contexts, deep neural networks are vulnerable to adversarial attacks. Typically, these attacks fool an image classification model by adding an unperceivable perturbation to the input image \cite{goodfellow2014explaining}. Despite the fact that the number of academic publications discussing end-to-end deep learning models is steadily increasing, their safety in real-world scenarios is still unclear. Though end-to-end models may lead to better performance and smaller systems, the monolithic module is also particularly vulnerable to adversarial attacks. In addition, note that current research on adversarial attacks primarily focuses on classification tasks. The effect of these attacks on regression tasks, however, largely remains unexplored. Our research explores the possibility of achieving real-time attacks against NVIDIA's end-to-end regression model (See Fig. \ref{fig:adv_drv}). However, the attacks may also be applied to the perception module in a modular driving system.

The main contributions of this paper are as follows: 

\begin{itemize}
    \item We propose two online white-box adversarial attacks against     an end-to-end regression model for autonomous driving: one strong attack that generates the perturbation for each frame (image-specific), and one stealth attack that produces a universal adversarial perturbation that attacks all frames (image-agnostic).
    \item The robustness of the attacks is illustrated using experiments conducted in Udacity Simulator. 
    The experiments demonstrate that it only takes the attack a few seconds to deviate the vehicle to outside of the lane.
    \item To facilitate future extensions and benchmark comparisons, our attack is open-sourced on Github\footnote{The code is available on Github: \url{https://github.com/wuhanstudio/adversarial-driving}}. As far as the authors are aware, this is the first open-source real-time attack on regressional driving models.
\end{itemize}

% \clearpage

\section{Preliminaries}

% Modular System

This section categorizes and describes end-to-end driving systems and associated adversarial attacks.

\subsection{End-to-End Driving Systems}

%Recent advances in autonomous driving simulators stimulate the development of end-to-end driving systems. Modular systems divide the driving pipeline into submodules and employ end-to-end models in the perception module. On the other hand, end-to-end driving systems treat the entire driving pipeline as a monolithic module that maps sensor inputs directly to steering commands \cite{Yurtsever2020}.

End-to-end driving systems treat the driving pipeline as a monolithic module that maps sensor inputs directly to steering commands \cite{Yurtsever2020}. Typically, end-to-end driving systems are implemented using either imitation learning or reinforcement learning. Imitation learning methods use deep neural networks to learn and mimic human driving behavior \cite{chen2022learning}. A supervisor is responsible for feeding the algorithm with labeled data. Reinforcement learning methods, on the other hand, improve driving policies via exploration and exploitation. The training process is not dependent on the existence of any supervisor. While there is a growing trend of publications that use reinforcement learning \cite{Chopra2020}\cite{perez2022deep}\cite{1286725}\cite{jaafra2019seeking}\cite{Chitta2021ICCV}, imitation learning is still more popular in end-to-end driving models \cite{tampuu2020survey}\cite{Prakash2021CVPR}\cite{Chitta2022PAMI}\cite{wu2022trajectory}. For this reason, our research will also focus on attacking imitation learning models.

The first implementation of an imitation-learning-based end-to-end driving system was the Autonomous Land Vehicle in a Neural Network (ALVINN) system, which trained a 3-layer fully connected network to steer a vehicle on public roads \cite{NIPS1988_812b4ba2}. However, end-to-end driving models have also been applied for the task of off-road driving \cite{NIPS2005_fdf1bc56}. More recently, researchers from NVIDIA built a convolutional neural network to map raw pixels from a single front-facing camera directly to steering commands \cite{bojarski2016end}. The NVIDIA end-to-end driving model is the target model in this paper, and details on this model are presented in Section \ref{section_problem_formulation}.

\subsection{Adversarial Attacks}
\label{adversarial_attacks}

This paper will consider an end-to-end driving model that outputs continuous steering commands, which is a regression model. Prior research on adversarial attacks primarily focuses on attacking classification models \cite{li2022review} \cite{zhang2021evaluating}\cite{boloor2019simple}\cite{abideen2022a3d}.

A successful attack against classification models deviates the output from the correct label. Taking the digital handwritten digit classification task as an example, an attacker can fool the classifier into recognizing the number 3 as 7. To evaluate the performance of an adversarial attack against a regression model, we need to quantify the magnitude of the resulting deviation. An attack that causes the steering angle to deviate from 1.00 to 0.99 will typically be considered unsuccessful since such a tiny deviation may not have any noticeable effect on the driving outcome. To be considered successful, an attack must lead to larger deviations. Prior research used Root Mean Squared Error (RMSE) \cite{msml} and Mean Square Error (MSE) \cite{nguyen2018adversarial} to evaluate and compare deviations. A successful attack should produce a higher MSE or RMSE than random attacks. Boloor et al. attacked an end-to-end self-driving model using human-perceivable physical shadow \cite{boloor2019}, while our research focuses on generating human-unperceivable perturbations.

While prior research primarily focuses on offline attacks against classification models, we investigate online attacks against regression models. Offline attacks apply perturbations to static images. Under the scenario of autonomous driving, an offline attack splits the driving record into static images and the corresponding steering angles. The perturbation is then applied to each static image, and the attack is evaluated using the overall success rate \cite{Deng2020}. Online attacks, on the other hand, apply the perturbation in a dynamic environment. Rather than applying the perturbation to static images in a driving record, we deploy the perturbation while the vehicle is driving. This also makes it possible to investigate the driving models' reactions to the attacks. 

One big difference between online and offline attacks is that the ground truth is unavailable in online attacks. Offline attacks take pre-recorded human drivers' steering angles as the ground truth, while real-time online attacks do not have access to pre-recorded human decisions. Therefore, we use the model output under normal benign conditions as the ground truth and assume that the driving model is comparatively close to the ground truth. This assumption is reasonable since if the model is inaccurate, the erroneous model is already a threat in itself. There is no need to attack the system in the first place.

Existing adversarial attacks can be categorized into white-box, gray-box, and black-box attacks \cite{REN2020346}: In white-box attacks, the adversaries have full knowledge of the target model, including model architecture and parameters; In gray-box attacks, the adversaries have partial information about the target model; In black-box attacks, the adversaries can only gather information about the model through querying. Since white-box attacks are more efficient, we devise two white-box attacks that achieve real-time performance against end-to-end driving models .

\clearpage

\section{Problem Formulation}
\label{section_problem_formulation}

In this section, we specify our objective, introduce mathematical notation, and describe our target model. Throughout the paper, we will use the notation
\begin{align}
y&=f(\theta, x) \\
y'&=f(\theta, x')
\end{align}
where $y$ is the benign output steering command, $f(\theta, x)$ is the regression model that maps input images to steering commands, $\theta$ is the model parameters, $x$ is the original input image, $y'$ is the adversarial output steering command, and $x'$ is the adversarial input image. Further, we will use $\eta=x-x'$ to denote the adversarial perturbation, $y^{*}$ is the ground truth steering command, and $J(y, y^{*})$ to denote the training loss. Given an input image $x$, the objective of attacking a classifier is to generate a small perturbation $\eta$, such that $y' \neq y^{*}$. However, the objective of attacking a regression model is to generate a small perturbation $\eta$, such that the difference between $y'$ and $y^{*}$ is larger than the average deviation caused by adding random noise to $x$. 

We use the $L_2$ norm to quantify the magnitude of the perturbation. The $L_2$ norm of the perturbation $\eta$ should be smaller than 0.03 (8 / 255) for an RGB input image according to the value used in prior research \cite{chow2020adversarial} \cite{ACFH2020square}. In particular, to ensure that the perturbation is unperceivable to human eyes, we require \begin{equation}
||x^{'}-x||_{2} = ||\ {\eta}\ ||_{2} \leq \xi
\end{equation}
where $\xi = 0.03$.

Our target model is the NVIDIA end-to-end driving model \cite{bojarski2016end}. The input shape of the model is (160, 320, 3), which represents (height, width, channel) respectively. The output steering angle is in the range of $[-1, 1]$ on all our (unperturbed) collected images. An output of $-1$ represents steering to the left, and an output of 1 means steering to the right. The input image is captured by the front camera, and we then apply predefined preprocessing methods before feeding the image to the model. Refer to \cite{bojarski2016end} for details on these preprocessing methods, including cropping, resizing, and RGB to YUV.

\section{Adversarial Attacks} 

In this section, we devise two white-box attacks against the driving system: one image-specific attack and one image-agnostic attack. Then, we present the system architecture.

\subsection{Image-specific Attack}

The first adversarial attack against a classifier was an image-specific offline attack that generated one perturbation for every input image \cite{goodfellow2014explaining}. Instead of minimizing the training loss $J(y,\ y^{*})$, Goodfellow et al. maximized the training loss and then used the gradient of the training loss to generate the perturbation. However, online attacks do not have access to the ground truth $y^{*}$, and thus, for online attacks, the training loss $J(y, y^{*})$ cannot be calculated. As a result, we need a new adversarial loss $J(y)$ that only requires the model output $y$ to generate the perturbation.

When attacking a regression model, notice that we have the choice to either increase or decrease the output. For example, to attack the end-to-end driving regression model, we can either deviate the vehicle to the left by decreasing the output or to the right by increasing the output. Therefore, in some sense, attacks on regression models can be seen as a special case of attacks on classification models, with the constraint that we only have two choices: increasing or decreasing the output. Accordingly, we will consider the straightforward adversarial loss functions 
% \begin{equation}
% \eta=\epsilon \text{sign}(\nabla_{x}J(\theta, x, y)).
% \end{equation}
\begin{align}
    J_{\texttt{left}}(y) &= -y \\
    J_{\texttt{right}}(y) &= y
\end{align}
for the image-specific attack. 

As explained in Section \ref{adversarial_attacks}, the adversarial loss functions $J(y)$ do not include ground truth $y^{*}$, which we do not have access to for online attacks. We can then utilize the Fast Gradient Sign Method (FGSM) to generate perturbations as 
\begin{equation}
    \eta = \epsilon \ \text{sign}[\nabla_{x}( J(y))]
\end{equation}
where $\epsilon$ is a scaling factor that determines the visibility of the perturbation. The image-specific attack is summarized in Algorithm \ref{alg:image-specific}.

As an example, assume that the attacker wishes to attack the vehicle to the right side. In this case, the objective is to increase the model output. We can then use the adversarial loss $J_{right}(y)$ to generate the perturbation. $\nabla_{x}( J(y))$ represents the gradient of the adversarial loss over the input. The gradient gives us information regarding how changes in the adversarial loss $y$ will back-propagate to the input.

%Our experimental results show that the image-specific attack is a strong attack that deviates the vehicle to the outside of the lane in several seconds.

\begin{algorithm}[t]
    \caption{Image-specific Attack}\label{alg:image-specific}
    \begin{algorithmic}
        \State Input: The regression model $f(\theta, x)$, the input images $\{x_t\}$ where $x_t$ is the image at time step $t$.
        \State Parameters: The strength of the attack $\epsilon$.
        \State Output: Image-specific perturbation $\eta$.
        \For{each time step $t$}
            \State Inference: $y = f(\theta, x)$
            \State Perturbation: $\eta = \epsilon \ \text{sign}[\nabla_{x}( J(y))]$
        \EndFor
    \end{algorithmic}
\end{algorithm}

\begin{algorithm}[t]
    \caption{Image-agnostic Attack (Training)}\label{alg:image-agnostic}
    \begin{algorithmic}
        \State Input: The regression model $f(\theta, x)$, input images in a driving record $X$, the target direction $I \in \{-1, 1\}$.
        \State Parameters: the number of iterations $n$, the learning rate $\alpha$, the step size $\xi$ , and the strength of the attack $\epsilon$ measured by the $l_{\infty}$ norm.
        \State Output: Image-agnostic perturbation $\eta$.

        \State Initialization: $\eta \leftarrow 0$
        \For{each iteration}
            \For{each input image $x$ in the driving record $X$}
                \State Inference: $y = f(\theta, x + \eta)$
                \If {$\text{sign}(y) \neq I$}
                    \State $x^{'} = x + \eta$
                    \State $\eta_{t} \leftarrow 0$
                    \While{$\text{sign}(y) \neq I$}
                        \State Gradients: $\nabla = \frac{\partial J(y)}{\partial x'}$
                        \State Perturbation: $\eta_{t} = \eta_{t} + proj_{2}(\nabla,\ \xi)$
                        \State Inference: $y = f(\theta, x + \eta_t)$
                    \EndWhile
                    \State $\eta = proj_{\infty}(\eta + \frac{\alpha}{\xi} \eta_{t},\ \epsilon)$
                \EndIf
            \EndFor
        \EndFor
    \end{algorithmic}
\end{algorithm}

\subsection{Image-agnostic Attack}

\begin{figure*}[b]
    \centering
    \includegraphics[width=0.75\textwidth]{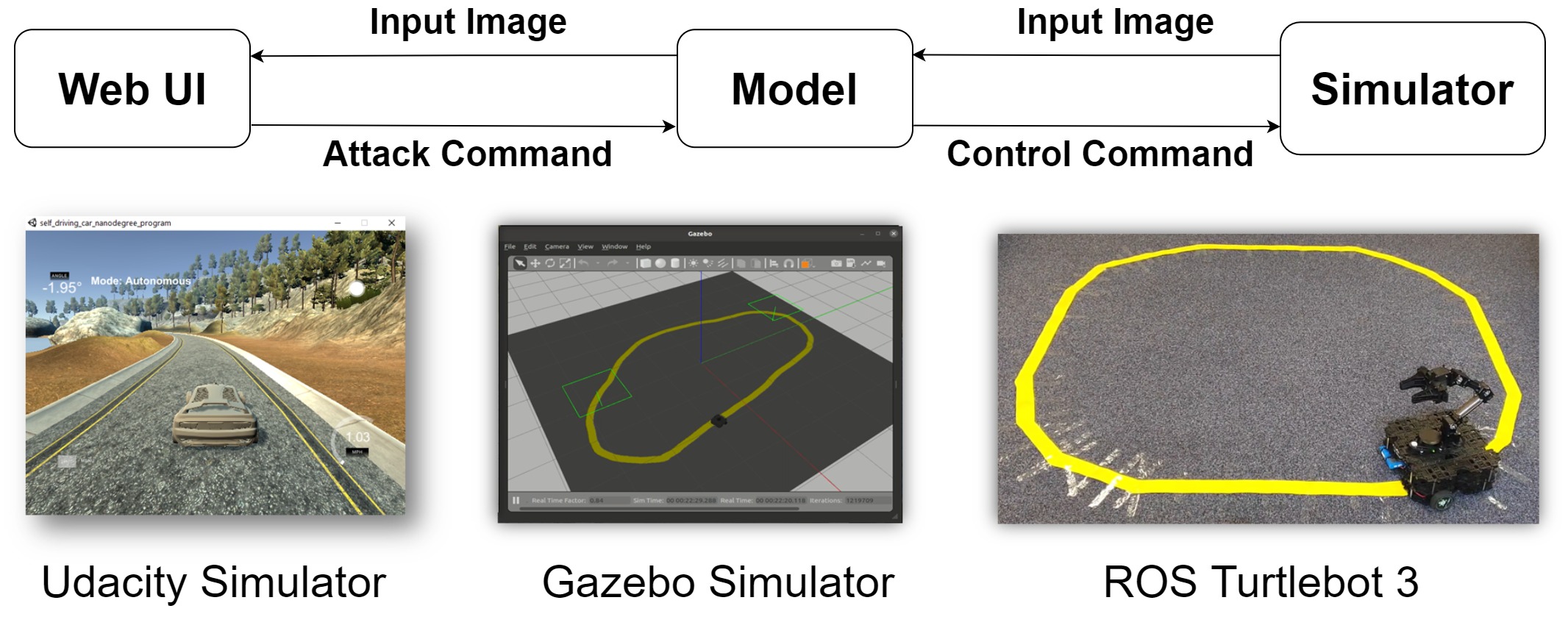}
    \caption{The architecture of the Adversarial Driving System. We tested our attacks in three environments: the Udacity Simulator, the ROS Gazebo Simulator, and a real Turtlebot 3.}
    \label{fig:arch}
\end{figure*}

Even small deviations may cause traffic accidents. A small deviation in the steering angle may, for example, result in a failure to steer around a sharp corner. In other words, even if the attack is not as strong as the image-specific attack it could still be perilous if applied at critical time points. Therefore, we introduce a white-box attack that generates a universal adversarial perturbation (UAP) \cite{moosavi2017universal} which can be used to attack all input images at different time steps. The image-agnostic attack combines the idea of DeepFool \cite{moosavi2016deepfool} and Projected Gradient Descent (PGD) \cite{madry2017towards}. The attack consists of two procedures: training and deployment. We first generate a UAP online or via a driving record and then deploy the UAP.

We first decide our target direction, that is, whether to attack the vehicle to the left ($y < 0$) or to the right ($y > 0$), and then choose the corresponding adversarial loss function ($J_{\texttt{left}}(y)$ or $J_{\texttt{right}}(y)$). The perturbation is initialized as zero. For each input image at each timestep, if the direction of the model output is not the same as the desired direction, we find the minimum perturbation that changes the sign of the model output to the desired direction. 

To change the direction of the model output with the minimum perturbation, we calculate the gradient of the adversarial loss $J(y)$ and then project the gradient to the $L_2$ ball. The closed-form solution to the optimization problem $\arg \min \ ||\ \eta - \eta^{'} \ ||_{2}$ with the constraint $||\ \eta^{'} \ || \leq \xi$ is given by
\begin{equation}
    proj_{2}(\eta,\  \xi) = \frac{\eta}{\max\{1, \frac{||\ \eta\ ||}{\xi}\}} =  \eta \min\{1, \frac{\xi}{||\ \eta\ ||} \}
\end{equation}
which can be proved using the Lagrangian and the KKT conditions \cite{boyd2004convex}.

% \begin{equation}
% \arg\underset{{\eta}^{'}}{\min }||\eta-\eta^{'}||_{2} \text{ subject to } ||{\eta}^{'}||_{2} \leq \xi
% \end{equation}

% \begin{equation}
% \Delta \eta_{t} = proj_{2}(\eta,  \xi) = \xi \frac{\nabla_{x} [J(y)]}{||\ \nabla_{x} [J(y)]\ ||_{2}}
% \end{equation}

After applying the temporary perturbation $\eta_{t}$ at timestep $t$, if the direction of the model output matches the desired direction, we incorporate the temporary perturbation $\eta_{t}$ to the overall perturbation $\eta$ and then project $\eta$ on the $l_2$ ball centred at 0 and of radius $\epsilon$ to ensure that the constraint $||{\eta}^{'}||_{2} \leq \epsilon$ is satisfied. The attack is summarized in Algorithm \ref{alg:image-agnostic}. As can be seen, the attack uses a similar while loop as in DeepFool and the projection function introduced in the PGD attack.

%The strength of the attack is not as strong as the image-specific attack, but it still adds an opposite force to the vehicle while turning through the corner.

\subsection{System Architecture}

The Robot Operating System (ROS) \cite{ros} is the most popular software framework in robotic research and applications. One example of an attack that
injects malicious data into a running ROS application is the Stealth Publisher Attack \cite{dieber2020penetration}. We exploit the same vulnerability to inject adversarial perturbations into a running end-to-end driving ROS application.

We design an adversarial system to attack the end-to-end autonomous driving system (See Fig. \ref{fig:arch}). The system consists of three key components: the simulator, the server, and the Web User Interface (UI). The simulator publishes the image captured by the front camera to the server. Meanwhile, it accepts steering commands from the server to manipulate the vehicle. The modular design pattern makes it possible to conveniently replace the simulator with a real Turtlebot without breaking the whole system. The server receives input images from the simulator via WebSocket connections and then sends back the control commands. Meanwhile, it receives attack commands from the web browser and then injects the adversarial perturbation into the input image. The end-to-end driving model is deployed on the server as well. We use a website as a front-end where the attacker can monitor the status of the simulator and choose different attacks.
The experimental results are presented in the next section.

\clearpage

\section{Experimental Results}

This section first describes the training of the driving systems. Following this, we describe the performance of our proposed image-specific and image-agnostic attacks.

% \begin{figure*}[b]
%     \centering
%     \includegraphics[width=\textwidth]{figures/ros_turtlebot.jpg}
%     \caption{Our test platforms and input images in ROS: A real turtlebot and ROS Gazebo Simulator.}
%     \label{fig:turtlebot}
% \end{figure*}

%\vspace{-1ex}

\subsection{Model Training}

Our objective is to implement a real-time online attack against an end-to-end imitation learning model. Since it is risky to perform online attacks against real-world driving systems, we tested our attacks in self-driving simulators. 

The target imitation learning models were trained from human driving records. In total, we collected 32k images of human driving records in our test environments: the Udacity Simulator (8k), the Gazebo Simulator (12k), and a real Turtlebot 3 (12k). We then trained one end-to-end driving model for each individual environment. The structure of the driving model is detailed in Table \ref{table_NVIDIA_model}.
Our experiments showed that all three models were vulnerable to adversarial attacks. 

In the following sections, we use the data from the Udacity Simulator to analyze the attack since experiments in this environment are easier to reproduce and examine than using a Turtlebot for other researchers. The experiments conducted using the Gazebo simulator are illustrated in the demo video.

% Fig. \ref{fig:turtlebot} illustrates the input image we collected in Gazebo Simulator and from a real Turtlebot 

\begin{table}[H]
\begin{center}
\begin{tabular}{ c c c }
     \hline
     Layer & Output Shape & Parameters \\ 
     \hline
     Input & (None, 160, 320, 3) & 0 \\  
     Conv2D & (None, 78, 158, 24) & 1824 \\  
     Conv2D & (None, 37, 77, 36) & 21636 \\  
     Conv2D & (None, 17, 37, 48) & 43248 \\  
     Conv2D & (None, 15, 35, 64) & 27712 \\  
     Conv2D & (None, 13, 13, 24) & 36928 \\
     Dropout & (None, 13, 13, 24) & 0 \\
     Flatten & (None, 27456) & 0 \\
     Dense & (None, 100) & 2745700 \\
     Dense & (None, 50) & 5050 \\
     Dense & (None, 10) & 510 \\
     Dense & (None, 1) & 11 \\
     \hline
    \end{tabular}
    \caption{The structure of the end-to-end driving model. \label{table_NVIDIA_model}}
\end{center}
\end{table}

\subsection{The Image-Specific Attack}

To begin with, we demonstrate that applying random noise to the end-to-end driving model only results in very small deviations. The parameter $\epsilon$ is used to ensure that the total disturbance of the random noise is the same as from the image-specific attack. Specifically, note that the image-specific attack adds or subtracts $\epsilon$ from each pixel based on the sign of the gradient. Likewise, we construct a random noise perturbation that randomly adds or subtracts $\epsilon$ from each pixel. 

\begin{figure}[tbph]
    \centering
    \begin{subfigure}[b]{0.47\textwidth}
        \centering
        \includegraphics[width=\textwidth]{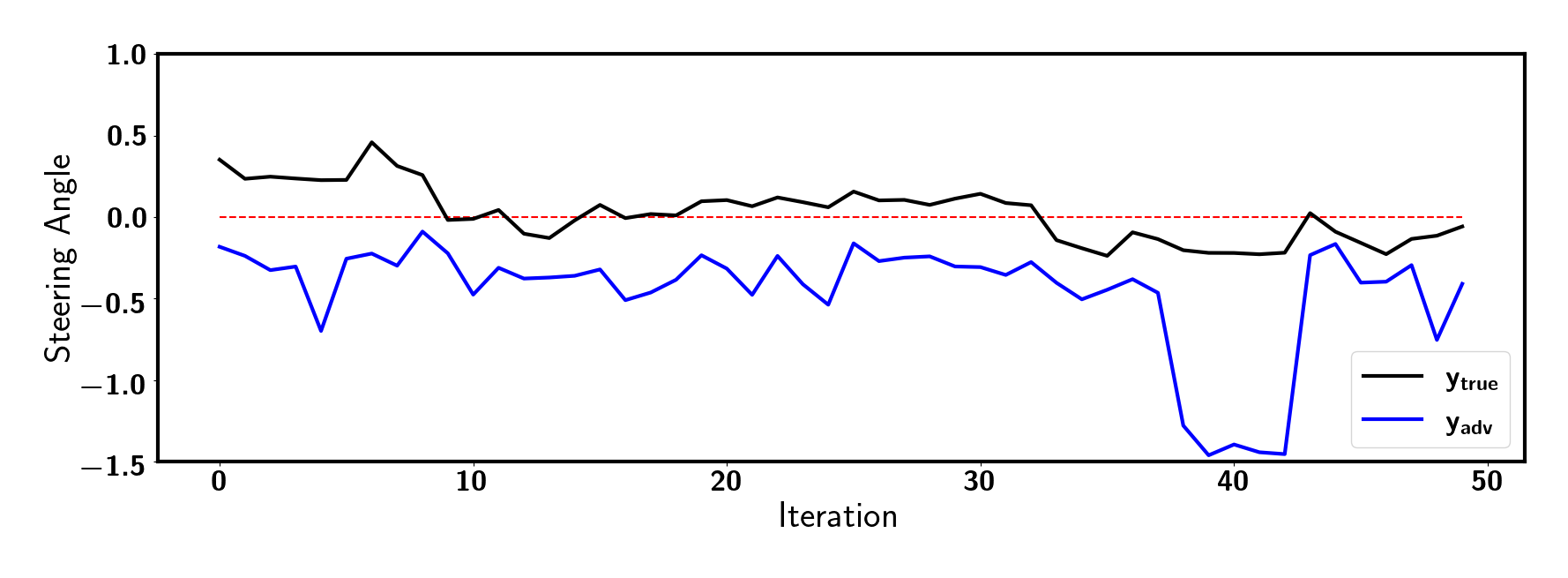}
        \caption{The image-specific left attack decreases the steering angle.\label{fig:image-specific-left-attack}}
    \end{subfigure}
    \begin{subfigure}[b]{0.47\textwidth}
        \centering
        \includegraphics[width=\textwidth]{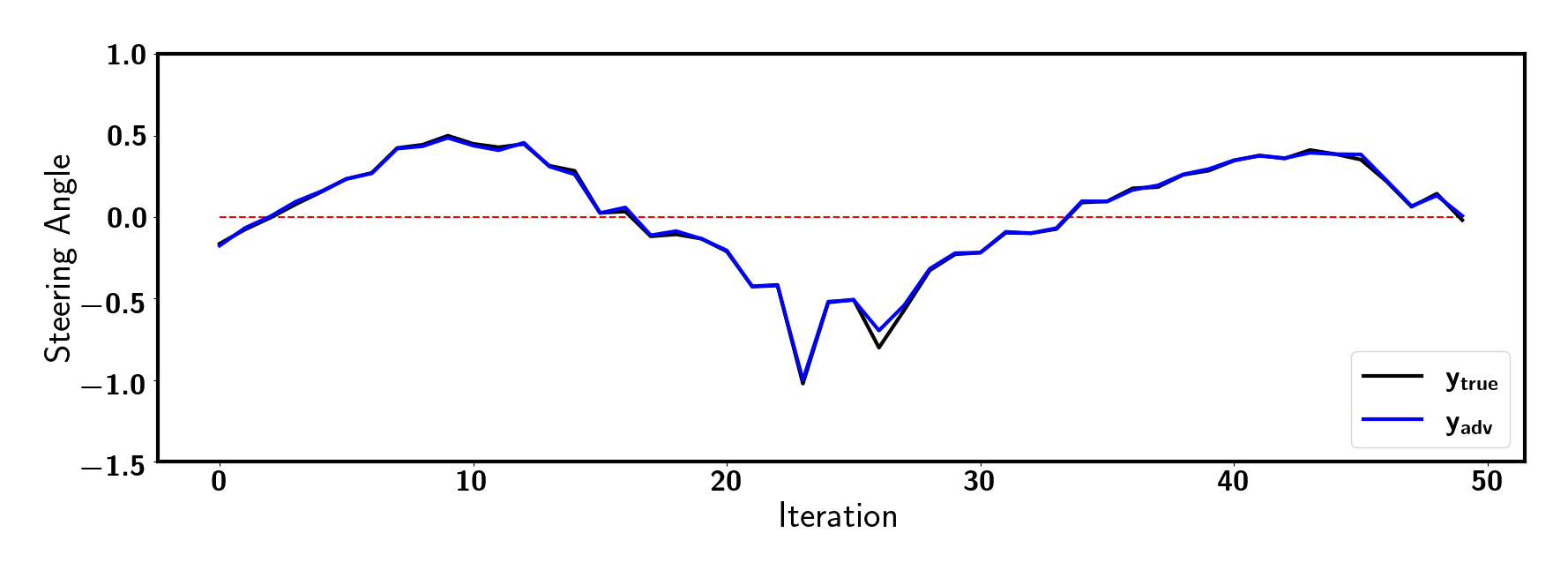}
        \caption{The random noise perturbation barely deviates $y_{adv}$ from $y_{true}$.}
    \end{subfigure}
    \begin{subfigure}[b]{0.47\textwidth}
        \centering
        \includegraphics[width=\textwidth]{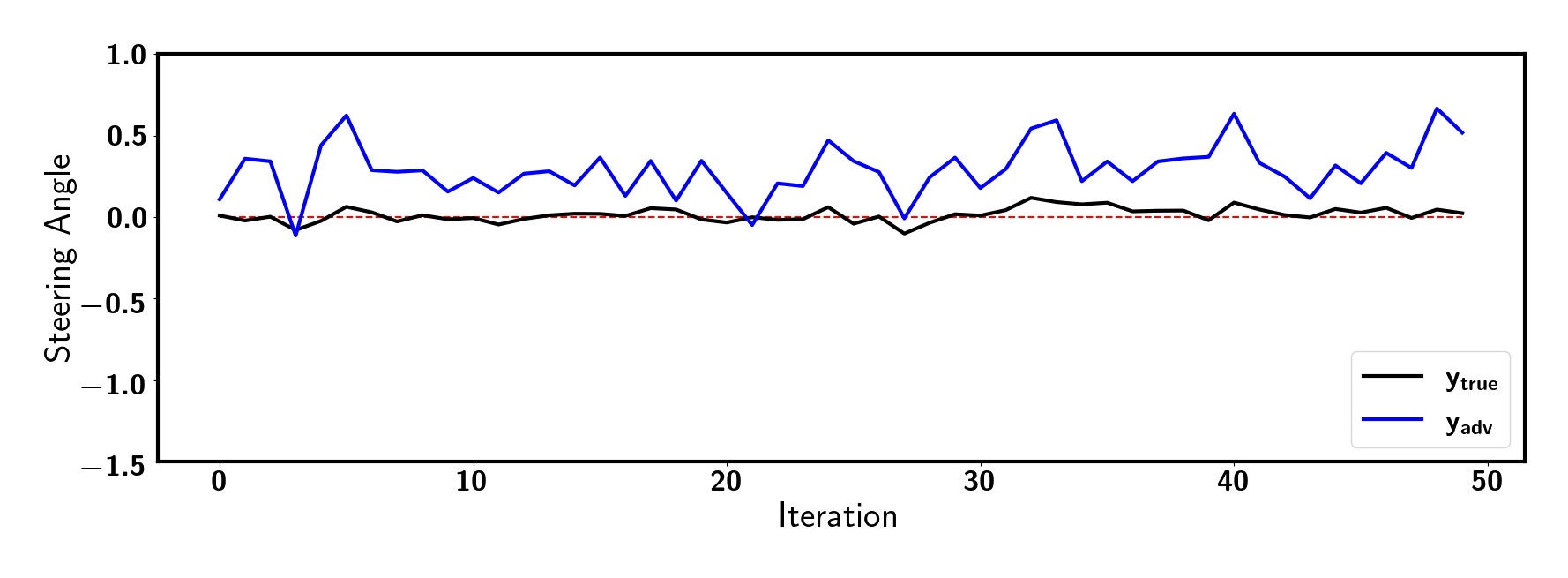}
        \caption{The image-specific right attack increases the steering angle.\label{fig:image-specific-right-attack}}
    \end{subfigure}
    \caption{The image-specific attack and random noise with the same strength ($\epsilon = 1$).}
    \label{fig:image-specific-online}
\end{figure}

In Fig. \ref{fig:image-specific-online}, we applied three different attacks that are of the same strength. Once under the image-specific attack, the vehicle drove off the road in several seconds. The image-specific left attack deviates the vehicle to the left by decreasing the steering angle, thus the $y_{adv}$ is smaller than $y_{true}$ in Fig. \ref{fig:image-specific-left-attack}. On the other hand, the image-specific right attack deviates the vehicle to the right by increasing the steering angle, thus the $y_{adv}$ is greater than $y_{true}$ in Fig. \ref{fig:image-specific-right-attack}. The random noise perturbation barely deviates $y_{adv}$ from $y_{true}$, indicating that it has little effect on the driving model.

The image-specific attack achieved 20-30 FPS on an Intel i7-8665U CPU and 600-700 FPS on an NVIDIA RTX 2080Ti GPU. Since the CPU and GPU are also utilized for Udacity Simulator, the attack performance varies depending on the hardware temperature.

Further, we measured the MAD of the steering angle over 800 attacks. The results are shown in Table \ref{tab:image-specific}. As can be seen, even the weakest image-specific attack ($\epsilon=0.1$) is much stronger than the strongest random noise attack ($\epsilon=8$). When $\epsilon = 4$ and $\epsilon = 8$, we can even deviate the steering angle outside of the range $[-1, 1]$. In other words, the image-specific attack is very strong. However, its weakness is that it needs to calculate the gradients of each individual input image. In a real-world scenario, we may not have access to the input image and gradients. Thus, we propose the image-agnostic attack that trains the perturbation using driving records and does not need access to the input and gradients during the deployment.

\begin{table}[H]
    \centering
    \begin{tabular}{ccc}
    \hline
    Attack Strength & Random Noise Attack & Image-Specific Attack\\
    \hline
    \ $\epsilon=0.1$    & 0.0002    & 0.1448 \\
    \ $\epsilon=1$      & 0.0020    & 0.4779 \\
    \ $\epsilon=2$      & 0.0048    & 0.7329 \\
    \ $\epsilon=4$      & 0.0150    & 1.4895 \\
    \ $\epsilon=8$      & 0.0278    & 2.4469 \\
    \hline
    \end{tabular}
    \caption{The mean absolute deviation of the steering angle over 800 image-specific attacks.}
    \label{tab:image-specific}
\end{table}

\pagebreak

\subsection{The Image-Agnostic Attack}

In similarity with the image-specific attack, the strength of the image-agnostic attack was also compared with a random noise attack.
The results are shown in Table \ref{tab:image-agnostic}. Though the image-agnostic attack is weaker than the image-specific attack, it is still stronger than the random noise attack.

\begin{table}[H]
    \centering
    \begin{tabular}{ccc}
    \hline
    Attack Strength & Random Noise Attack & Image-Agnostic Attack\\
    \hline
    \ $\epsilon=0.1$    & 0.0002    & 0.0373 \\
    \ $\epsilon=1$      & 0.0020    & 0.1109 \\
    \ $\epsilon=2$      & 0.0048    & 0.1294 \\
    \ $\epsilon=4$      & 0.0150    & 0.1131 \\
    \ $\epsilon=8$      & 0.0278    & 0.1275 \\
    \hline
    \end{tabular}
    \caption{The mean absolute deviation of the steering angle over 800 image-agnostic attacks ($\alpha=0.002,\ \xi=4$, $n=500$).}
    \label{tab:image-agnostic}
\end{table}

% We then tested different learning rate decay.

% Vheicles reaction to the attack

% As a result, the end-to-end driving model is vulnerable to adversarial attacks.

% \clearpage

\begin{figure}[t]
    \centering

    \begin{subfigure}[b]{0.48\textwidth}
        \includegraphics[width=\textwidth]{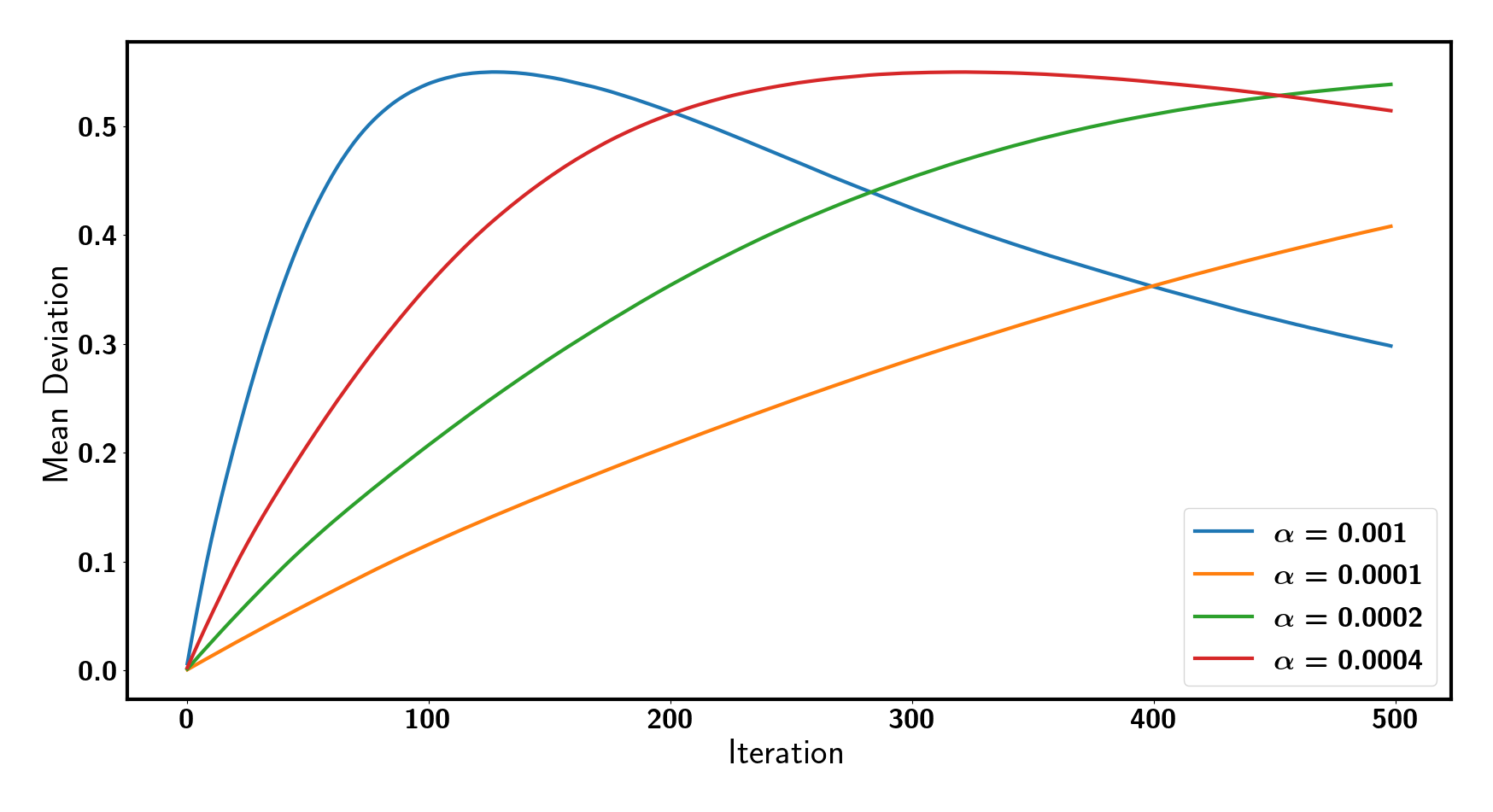}
        \caption{Different $\alpha$ with fixed $\epsilon=1, \xi=4$.}
        \label{fig:random}
    \end{subfigure}

    \begin{subfigure}[b]{0.48\textwidth}
        \includegraphics[width=\textwidth]{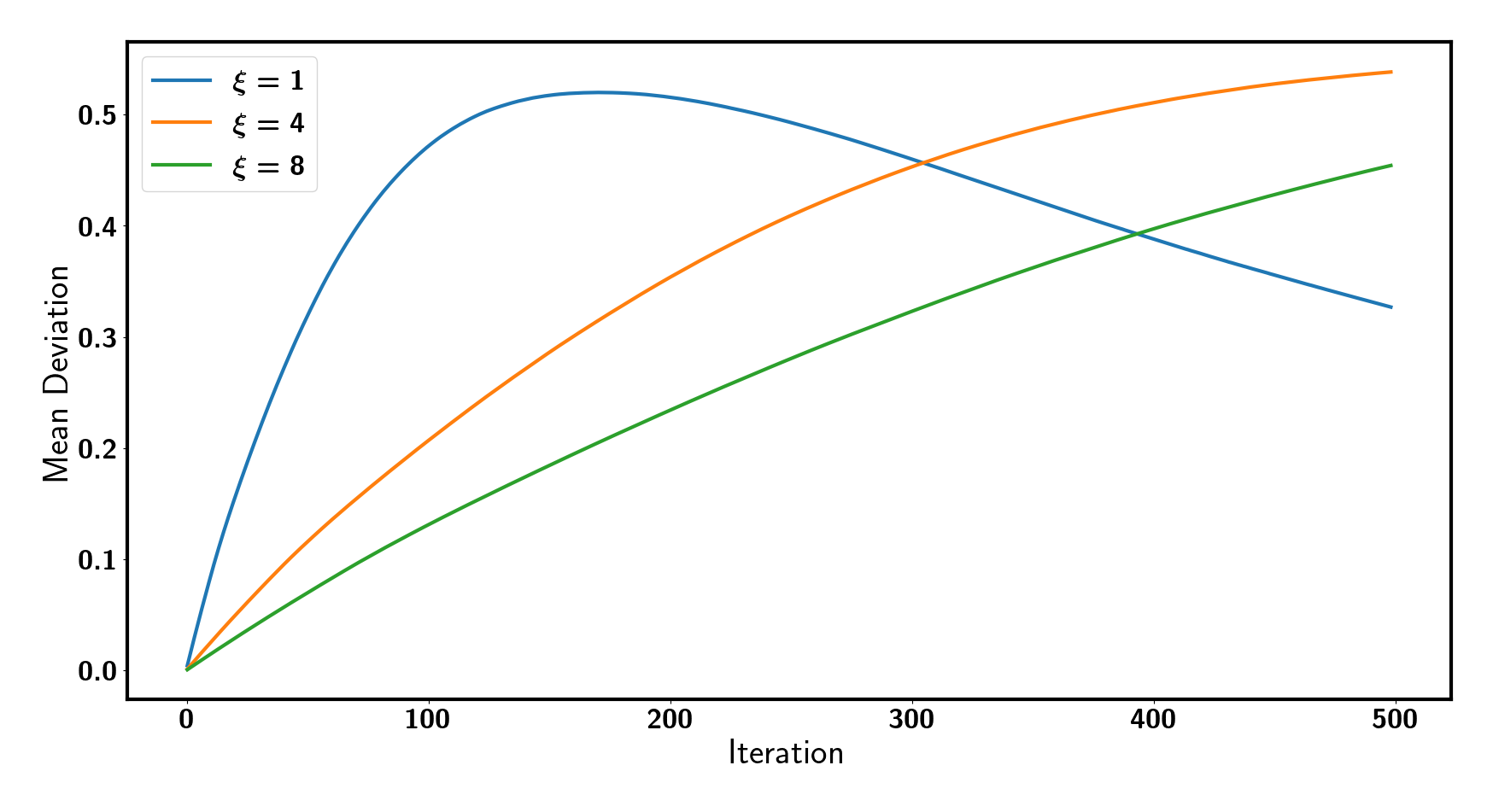}
        \caption{Different $\xi$ with fixed $\alpha=0.002$}
        \label{fig:left}
    \end{subfigure}

    \caption{The mean deviation of the steering angle during the training process with different hyperparameters.\label{figure_hyperparameters}}
\end{figure}

As seen in Table \ref{tab:image-agnostic}, the strength of the image-agnostic attack does not improve after $\epsilon > 2$. This is due to the limited generalizability of the perturbation. Increasing the strength of the attack further may increase the model output for some inputs but may equally well decrease the model for other inputs. Therefore, increasing $\epsilon$ further adds more variation to the model prediction while the MAD remains stable.

We also investigated the effect of the learning rate $\alpha$ and the step size $\xi$ on the training process (See Fig. \ref{figure_hyperparameters}). The learning rate $\alpha$ controls the variation of the perturbation during the whole iteration. We tested different $\alpha$ with fixed parameters $\epsilon=1$ and $\xi=4$. As $\alpha$ increases, the mean deviation increases faster. However, the iteration process also becomes more variable, and the mean deviation decreases after 100 steps when $\alpha>0.01$. 

The step size $\xi$ decides how fast the perturbation is updated to change the model output to the desired direction for each input image $x$. A smaller $\xi$ makes the update towards the target direction more steady, but the iteration takes a longer time. A larger $\xi$ can change the direction of the model output in a single step, but the perturbation may not generalize well to other inputs.

As illustrated in Fig. \ref{fig:image-agnostic-online}, using the parameters $\alpha=0.0002$ and $\xi=4$ enabled us to generate image-agnostic perturbations at $\epsilon=1$ that are comparable in performance with the image-agnostic attack at $\epsilon=0.1$. While the image-agnostic is not as strong as the image-specific attack, the image-agnostic attack makes the vehicle difficult to control at sharp corners (this is illustrated in the demo video), which could lead to incidents at some critical points. 

In addition, the image-agnostic attack applies the same perturbation to all frames. Thus, the deployment of the image-agnostic attack is much more computationally efficient than the image-specific attack.

% \pagebreak

\begin{figure}[tb]
    \centering
    \begin{subfigure}[b]{0.48\textwidth}
        \includegraphics[width=\textwidth]{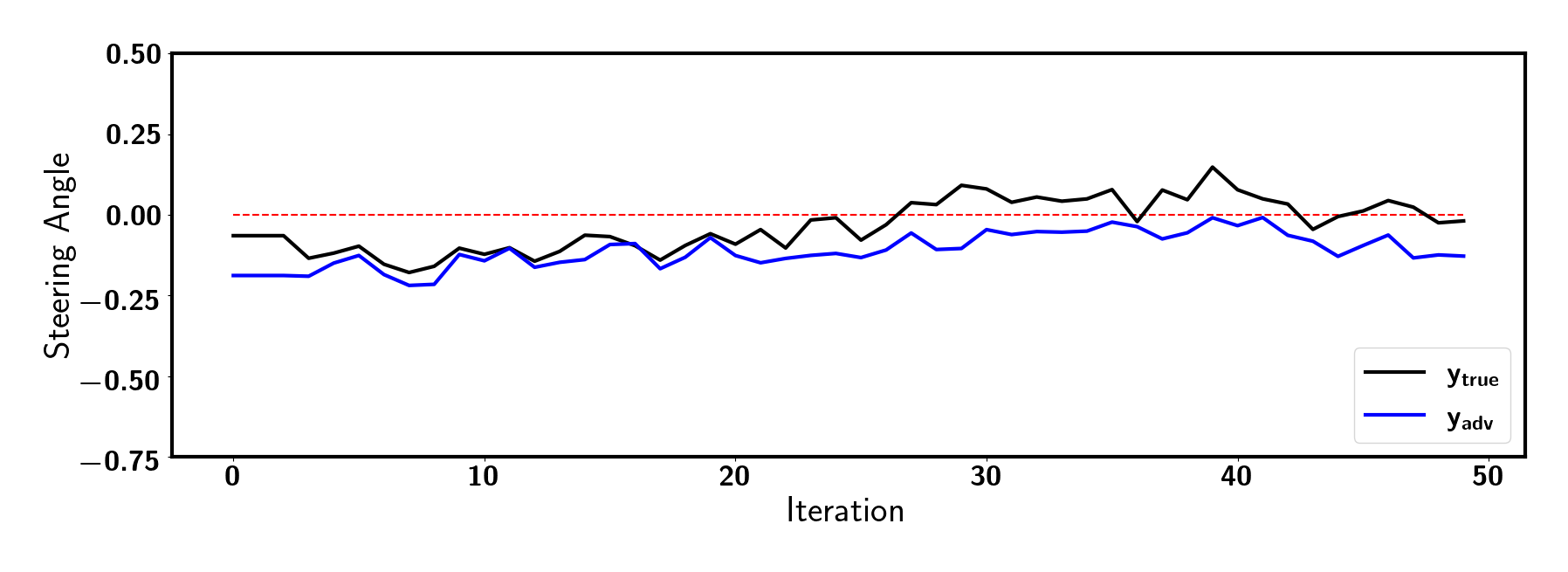}
        \caption{The image-agnostic Left Attack decreases the model output ($y_{adv}<0$), making it difficult to turn right. ($\epsilon=1)$}
        \label{fig:uni_left}
    \end{subfigure}
    \begin{subfigure}[b]{0.48\textwidth}
        \includegraphics[width=\textwidth]{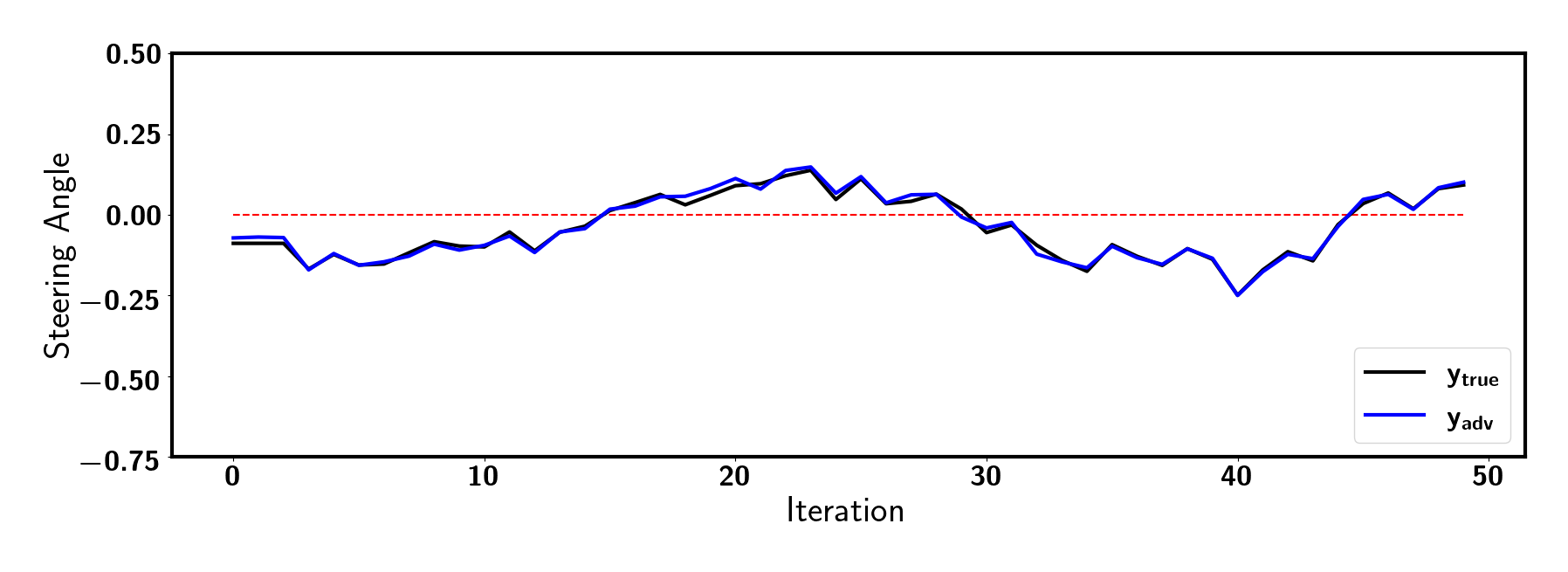}
        \caption{The random noises barely deviates $y_{adv}$ from $y_{true}$.}
        \label{fig:uni_random}
    \end{subfigure}
    \begin{subfigure}[b]{0.48\textwidth}
        \includegraphics[width=\textwidth]{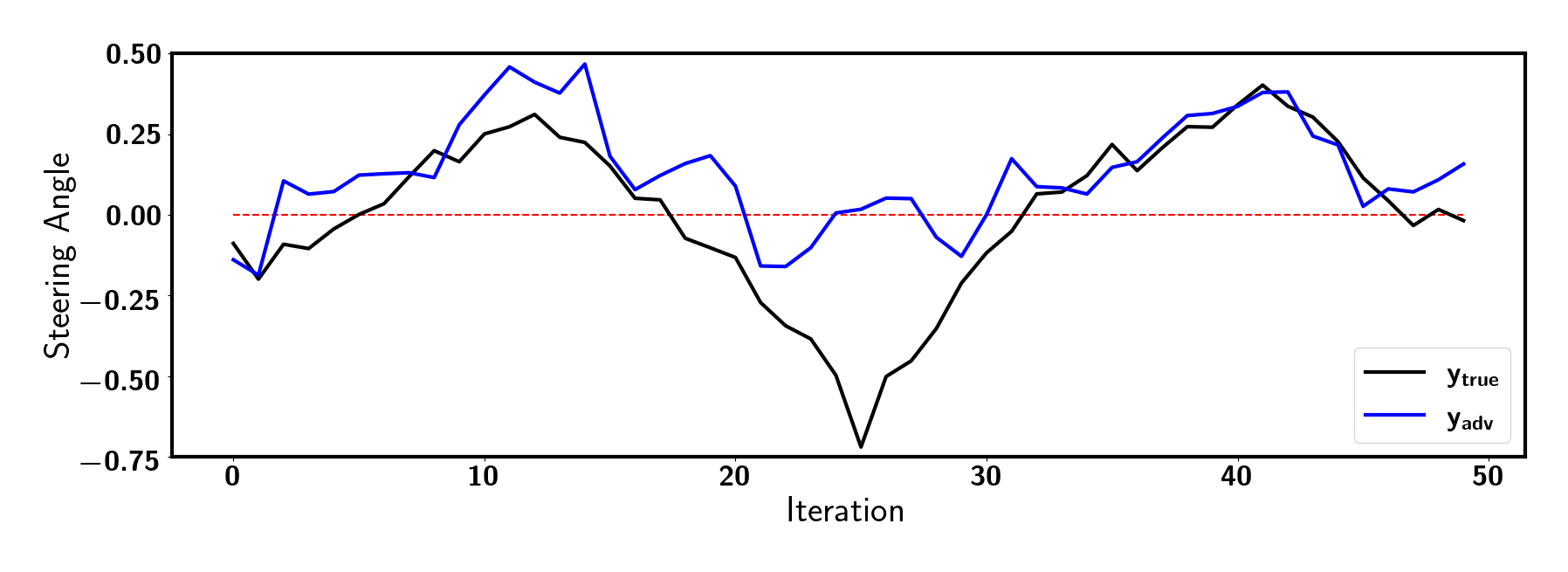}
        \caption{The image-agnostic Right Attack increases the model output ($y_{adv}>0$), making it difficult to turn left. ($\epsilon=1)$}
        \label{fig:uni_right}
    \end{subfigure}
    \caption{The Image-Agnostic attack ($\alpha=0.002$, $\xi=4$, $n=500$) and random noises with the same strength ($\epsilon = 1$).}
    \label{fig:image-agnostic-online}
\end{figure}

\addtolength{\textheight}{-2cm}

\clearpage

\section{Conclusions}

This paper has demonstrated that it is possible to attack an end-to-end driving model in real-time. We devise a strong image-specific attack and a stealthy image-agnostic attack. Though the mean absolute deviation of the image-agnostic attack is smaller than the image-specific attack, both attacks are more effective than random noise attacks. The image-agnostic attack deviates the vehicle outside of the lane after just a few seconds, while the image-agnostic attack could cause incidents at sharp corners. These results provide new evidence of the vulnerability of safety-critical robotic applications.

\bibliographystyle{IEEEtran}
\bibliography{IEEEabrv, mybibfile}

\end{document}